\documentclass{article} 
\usepackage{iclr2026_conference,times}

\usepackage{amsmath,amsfonts,bm}

\def\eqref#1{equation~\ref{#1}}

\def\1{\bm{1}}

\DeclareMathAlphabet{\mathsfit}{\encodingdefault}{\sfdefault}{m}{sl}
\SetMathAlphabet{\mathsfit}{bold}{\encodingdefault}{\sfdefault}{bx}{n}

\newcommand{\R}{\mathbb{R}}

\usepackage{hyperref}
\usepackage{url}
\usepackage{booktabs}
\usepackage{multirow}
\usepackage{xcolor}
\usepackage{makecell}
\usepackage{graphicx}
\usepackage{float} 
\title{Physics-informed fine-tuning of foundation models for partial differential equations}

\author{Vlad Medvedev, Leon Armbruster, Christopher Straub, Georg Kruse, \& Andreas Rosskopf 
\\
Department Modeling and Artificial Intelligence\\
Fraunhofer Institute for Integrated Systems and Device Technology IISB\\
Schottkystrasse 10, 91058 Erlangen, Germany \\
\texttt{\{vlad.medvedev, leon.armbruster, christopher.straub, georg.kruse,}\\
\texttt{andreas.rosskopf\}@iisb.fraunhofer.de}}
\iclrfinalcopy 
\begin{document}

\maketitle

    \begin{abstract} Foundation models for partial differential equations (PDEs) have emerged as powerful surrogates pre-trained on diverse physical systems, but adapting them to new downstream tasks remains challenging due to limited task-specific data and distribution shifts. While fine-tuning has proven transformative in natural language processing, best practices for adapting PDE foundation models remain underexplored. Although physics-informed training has successfully trained accurate solvers across a wide range of PDE problems, its potential for fine-tuning data-based foundation models has not been systematically studied. In this work, we introduce a physics-informed fine-tuning framework that adapts pre-trained PDE foundation models by incorporating physical constraints (PDE residuals and boundary conditions) directly into the fine-tuning objective. This enables effective adaptation in data-scarce regimes while promoting physical consistency. We evaluate our method on a downstream task composed of an unseen PDE class and compare it with data-driven finetuning counterparts. Our results demonstrate that physics-informed fine-tuning achieves competitive accuracy without requiring PDE solutions for training. Furthermore, a hybrid fine-tuning strategy yields superior generalization to out-of-distribution scenarios when only minimal training data is available. These findings establish physics-informed fine-tuning as a scalable and data-efficient paradigm, providing a physically interpretable pathway for adapting foundation models in scientific machine learning. \end{abstract}

\section{Introduction}
Partial differential equations (PDEs) serve as the mathematical foundation for a vast range of scientific and engineering disciplines, including climate modeling, energy systems, and aircraft design \citep{ye2025pdeformer, herde2024poseidon}. While numerical methods, such as finite difference and finite element methods, provide mathematically grounded, high-fidelity solutions, they often encounter prohibitive computational costs, particularly in many-query scenarios such as inverse problem solving and real-time control. Machine learning has emerged as a transformative alternative, providing differentiable surrogate models that can approximate complex physical systems with speedups of several orders of magnitude over conventional solvers~\citep{mccabe2025walrus}. 

The current standard for data-driven simulation surrogates relies on neural operators such as Deep Operator Networks~\citep{lu2019deeponet} and Fourier Neural Operators (FNOs)~\citep{li2020fourier}. These operators approximate the solution operator of a PDE, allowing them to map function space inputs (e.g., initial/boundary conditions or source terms) directly to the corresponding solution (trajectory). Despite their success, existing operator learning frameworks are domain-specific, meaning they are effective only within the narrow physical regimes for which they were trained. These models typically require retraining from scratch whenever the governing equations or physical parameters change, leading to significant computational overhead for retraining (and data acquisition).

To overcome these limitations, research is rapidly transitioning toward PDE foundation models: universal surrogates pre-trained at scale on diverse datasets spanning multiple physics. Models such as \textsc{Poseidon} \citep{herde2024poseidon}, \textsc{Walrus} \citep{mccabe2025walrus}, \textsc{PDEformer}-2 \citep{ye2025pdeformer}, \textsc{MORPH}\citep{rautela2025morph}, and \textsc{FM}-\textsc{PDE} \citep{soares2025towards} leverage shared representations across diverse physical systems and enable efficient adaptation to entirely new physical regimes, significantly reducing the data requirements for multi-physics simulation. By capturing common patterns across diverse PDEs, these foundation models can outperform standalone surrogates in data-scarce scenarios.

However, adapting a pre-trained model to unseen downstream tasks without extensive task-specific data remains challenging when task-specific data are limited.
This insufficiency becomes more pronounced for PDE classes with greater variety, where additional solution samples must be prepared using traditional numerical solvers to ensure accurate model predictions across the input domain.
Performance is also frequently hindered by distribution shifts in input functions: even when pre-training and downstream solutions exhibit similar physical behavior, differences in initial or boundary condition distributions prevent effective parameter transfer.
To address these challenges, researchers have explored distributed learning and robust averaged initializations, as well as data-efficient fine-tuning techniques \citep{zhang2025deeponet} that reduce computational burden, and physics-informed machine learning \citep{karniadakis2021,medvedev2025physics}, which alleviates data requirements by directly minimizing PDE residuals during training. While physics-informed fine-tuning has shown its advantages in multi-operator learning \citep{zhang2025deeponet}, its application to foundation models remains uninvestigated, representing an open research question.\par
In this paper, we address the gap between foundation model adaptation and physical fidelity by investigating physics-informed fine-tuning strategies for adapting pre-trained foundation models to downstream tasks involving unseen PDEs. Concretely, our contributions are:
\begin{itemize}
\item We introduce, for the first time, a physics-informed fine-tuning framework for data-based PDE foundation models. Specifically, we adapt pre-trained foundation models to new physical domains by incorporating PDE residuals and boundary conditions directly into the fine-tuning objective, thereby enabling effective adaptation in data-scarce or even zero-shot scenarios.
\item We demonstrate that hybrid fine-tuning (combining physics-informed and data-driven objectives) achieves superior extrapolation to downstream tasks and generalization to out-of-distribution scenarios, when only minimal training data is available.
\item Our experimental results establish that physics-informed fine-tuning enables data-free learning of unseen PDE families, achieving competitive accuracy compared to conventional data-driven approaches. This constitutes a promising paradigm that combines easy-to-obtain pre-training data with complex governing physics during fine-tuning, providing a scalable, data-efficient, and physically interpretable pathway for scientific machine learning.\end{itemize}

\section{Background}

\subsection{Neural operators for PDEs}
Neural operators represent a paradigm shift in PDE solving by learning operators between infinite-dimensional function spaces.
Consider a generic PDE problem on a domain $D\subset\R^d$ parametrized by an input $f\in\mathcal F$ (e.g., a source term) with the sought solution $u=u_f\in\mathcal U$:
\begin{equation}\label{eq:PDE_abstract}
    \mathcal N_f(u)=0\text{ on }D,\qquad \mathcal B_f(u)=0\text{ on }\partial D,
\end{equation}
where $\mathcal N_f$ denotes the differential operator and $\mathcal B_f$ the boundary condition operator.
The goal is to approximate the solution operator of Equation \ref{eq:PDE_abstract}, i.e., $\mathcal S\colon \mathcal F\to\mathcal U$ such that $\mathcal S(f)\approx u_f$. 
Hence, neural operators are surrogates specialized to the specific PDE problem they have been trained on.

The most prominent example is the {Fourier Neural Operator (FNO)} introduced by \cite{li2020fourier}, which is based on trainable kernels in Fourier space. 
In particular, FNOs are known to be universal approximators for the solutions of a large class of PDEs~\citep{kovachki2021}.

\subsection{Foundation models for PDEs}

Drawing inspiration from the success of Large Language Models, PDE foundation models transition from building specialized surrogates, like neural operators, to developing universal solvers capable of cross-domain transfer. 
PDE foundation models exploit large-scale pre-training on heterogeneous datasets encompassing diverse physical phenomena to learn a general prior over physical solutions. This leverages the shared mathematical structures, such as diffusion or advection, which allows knowledge learned from one set of phenomena to transfer across others.
Pre-training on a broad distribution of operators allows PDE foundation models to extract a physical latent space that captures these universal motifs, thereby significantly reducing the sample complexity required for downstream tasks.
Architecturally, PDE foundation models mostly rely on scalable Transformers as their backbone, which allow them to process multi-scale features with high fidelity.
Prominent examples include \textsc{Poseidon} \citep{herde2024poseidon} and the more recently proposed \textsc{Walrus} \citep{mccabe2025walrus}.

The pre-training of PDE foundation models is based on large datasets of numerical solutions to a variety of PDEs, followed by supervised fine-tuning that relies on labeled data from the target domain.
This pervasive reliance on labeled data presents a fundamental challenge: when the available target data is sparse or noisy, the fine-tuned model may predict statistically likely but physically incorrect and potentially even impossible outcomes.

\subsection{Physics-informed machine learning}

Physics-informed machine learning addresses the limitations of purely data-driven models by embedding the governing equations directly into the loss function used for the training of the model~\citep{karniadakis2021}.
Specifically, for the PDE problem in Equation \ref{eq:PDE_abstract}, the physics-informed loss function $\mathcal L_{\text{physics}}$ is composed of the residuals of PDE $\|\mathcal N_f (u)\|$ and boundary condition $\|\mathcal B_f (u)\|$ evaluated at collocation points in $D$ and $\partial D$, respectively. To ensure generalization across~$f\in\mathcal F$, the loss is assessed for varying~$f$ during the training of a neural operator.
The derivatives contained in the differential operator~$\mathcal N_f$ can be computed via automatic differentiation or, in the case of a discretized model output, via a finite difference approximation.

While models trained in a physics-informed way do not require any data for training and excel at physically consistent extrapolation, they often suffer from a more unstable training.
Previous works have demonstrated that this can be mitigated by combining data and physics during training~\citep{zhang2025deeponet}.

\subsection{Benchmark PDE}\label{ssc:benchmark}
The two-dimensional Poisson equation is a fundamental elliptic PDE that appears across numerous scientific domains, including electrostatics, heat conduction, and potential theory. It is of the form
\begin{equation}
-\Delta u = f, \quad \text{ on } D:=(0,1)^2,
\label{eq:poisson}
\end{equation}
and is equipped with vanishing Dirichlet boundary condition $u = 0$ on $\partial D$. 
The source term $f$ can be generated as a superposition of a random number of Gaussian functions \citep{herde2024poseidon}. This construction ensures diverse input distributions while maintaining analytical tractability, making it well-suited for operator learning methods.

\section{Methodology}
\begin{figure}[ht]
\begin{center}
\includegraphics[width=\textwidth]{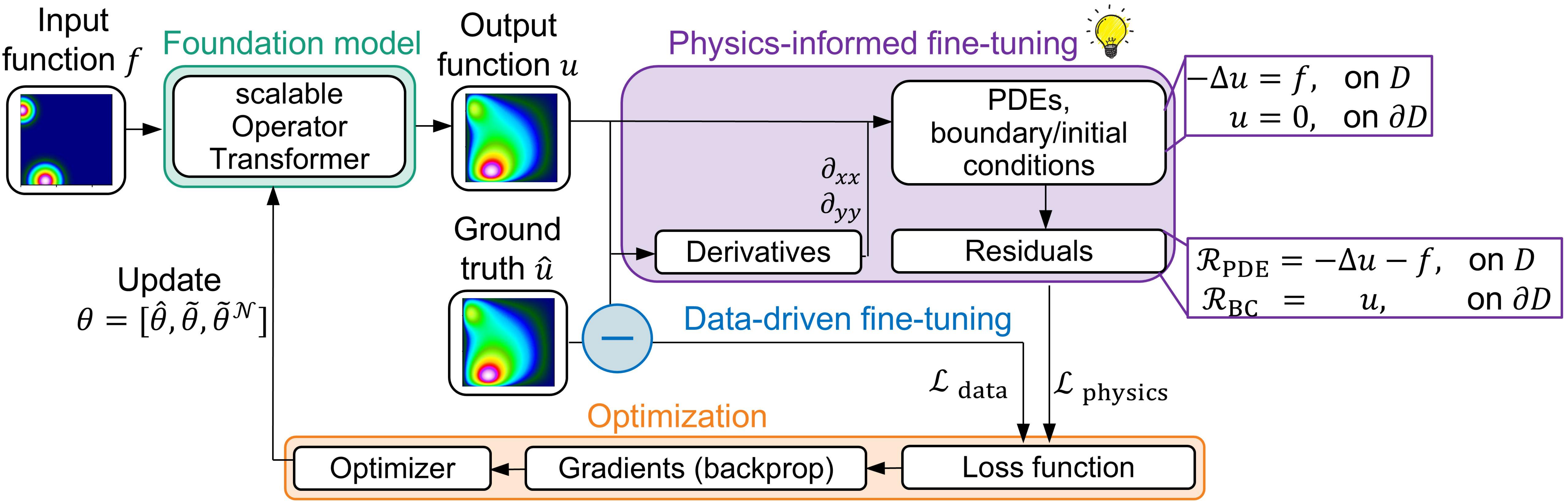}
\end{center}
\caption{Physics-informed fine-tuning of pre-trained foundation model with scalable Operator Transformer architecture \citep{herde2024poseidon} for the Poisson downstream task (cf.\ Section~\ref{ssc:benchmark}).}
\label{fig:figure1}
\end{figure}

We now present our approach to physics-informed fine-tuning of foundation models, as illustrated in Figure~\ref{fig:figure1}. Following the \textsc{Poseidon} framework \citep{herde2024poseidon}, we apply it here to the two-dimensional steady-state Poisson equation with general source as our sample downstream task to evaluate fine-tuning strategies. While the Poisson problem is relatively simple compared to complex multiphysics simulations, it serves as an effective proof of concept for several reasons. Firstly, it allows us to isolate the benefits of physics-informed fine-tuning from the complexities introduced by time-dependent dynamics. Secondly, the well-understood mathematical properties of the Poisson equation enable clear interpretation of model behavior and failure modes. Thirdly, this downstream task represents an out-of-distribution challenge, as both the initial data distribution and the governing physics differ fundamentally from pre-training. The time-independent, elliptic Poisson equation is dominated by diffusion and smoothing, contrasting sharply with the transport, shock propagation, and fluid mixing dynamics encountered during pre-training of \textsc{Poseidon}. Finally, the computational efficiency of generating ground truth data permits comprehensive studies and analysis. To demonstrate that our conclusions are not limited to a single PDE, we additionally consider the Helmholtz equation in Appendix \ref{sec:appendixB}.

The \textsc{Poseidon} training process involves large-scale pretraining on a diverse dataset of fluid dynamics governing equations using a novel \textsc{all2all} strategy that leverages the semi-group property of PDEs to quadratically scale training samples, followed by finetuning on a small number of task-specific examples to generalize to unseen and unrelated physical regimes. To adapt the \textsc{Poseidon} foundation model to our downstream task, we follow the fine-tuning protocol established by \citet{herde2024poseidon}, but adopt a physics-informed objective. Rather than minimizing only data-driven loss based on mismatch with pre-generated ground truth solutions, we additionally minimize the PDE residual, steering the solution toward physically consistent results. 
The model output is zero-padded at the domain boundaries to serve two purposes. Firstly, padding preserves spatial dimensions when computing PDE derivatives via finite differences, ensuring dimensional consistency between input and output. Secondly, zero-padding directly enforces zero Dirichlet boundary conditions (Equation~\ref{eq:poisson}), eliminating the typical need for soft penalty terms in the loss function \citep{karniadakis2021}. For other boundary conditions, the framework can incorporate alternatives, such as wrap-around padding for periodic boundaries. Finally, we minimize the physics-informed loss, defined as the mean squared error of the Poisson equation residual evaluated within the simulation domain $D$. When following a hybrid training approach, the standard data loss is added to the total loss. The gradients of this loss are backpropagated through the network to update the foundation model's weights via gradient-based optimization.\par
\textsc{Poseidon}'s learnable parameters $\theta$ are decomposed into three distinct subsets: the core backbone latent parameters $\hat{\theta}$, the embedding and recovery parameters $\tilde{\theta}$, and the lead-time conditioned layer-norm parameters $\theta_N$. For tasks involving PDE families not encountered during pre-training (e.g., previously unseen Navier–Stokes or Euler regimes), the backbone $\hat{\theta}$ and temporal embeddings $\tilde{\theta}_N$ are initialized via weight transfer from the pre-trained foundation model to leverage the rich physical representations acquired during the pre-training phase. Meanwhile, a third subset of weights $\tilde{\theta}$ is initialized from scratch with random weights rather than transferred. This replacement is necessitated by the fact that the number of input and output channels in downstream tasks frequently differs from the four-channel configuration utilized during the pre-training phase.\par
In the original work \citep{herde2024poseidon}, $\theta$ parameter subsets are updated with distinct learning rates during fine-tuning: while the core backbone is updated conservatively to preserve general knowledge, the embedding, recovery, and timing layers are updated more aggressively to handle the specifics of the new downstream physics. 
During the fine-tuning process, we maintain the trainability of the entire parameter set without freezing the latent space, thereby allowing the model to fully specialize to the target physics of the downstream PDEs.  Furthermore, steady-state PDEs (e.g., the Poisson equation) are handled similarly to \textsc{Poseidon} \citep{herde2024poseidon}, by interpreting them as the long-time limit of an evolutionary PDE.\par


\section{Experimental Setup}

\subsection{Datasets}
\label{ssc:datasets}
\paragraph{Dataset and Evaluation Protocol (Interpolation).}
We utilize the pre-generated datasets from the \textsc{Poseidon} framework \citep{herde2024poseidon} to ensure fair comparison with existing baselines. Following the established protocol, we reserve the first 240 samples exclusively for testing. 
To evaluate model performance across varying data regimes, we construct data scaling curves by training on labeled data subsets of size $M \in \{1, 2, 4, 8, 16, 32, 64, 128, 256, 512, 1024, 2048, 4096\}$. In the \textsc{Poseidon} dataset, the source term $f$ is generated as a superposition of randomly positioned Gaussian blobs with varying amplitudes and widths, as shown in Figure \ref{fig:figure2}. Since both training and test samples follow this distribution, these experiments primarily assess the \textit{interpolation} capabilities of the models under data-scarce conditions.

\paragraph{Out-of-Distribution Generalization (Extrapolation).}
To assess robustness against distribution shift, we generate an additional test set of 50 samples featuring extreme source shapes qualitatively different from the Gaussian blob shapes seen during training. These include sinusoidal waves, radial patterns, angular patterns, checkerboard patterns, random polygons, Perlin-like noise, stripe patterns, spiral patterns, and random lines (Figure \ref{fig:figure4}). 
Training on Gaussian sources while testing on these extreme shapes effectively probes the \textit{extrapolation} performance of the models, extending the scope of analysis beyond what was conducted in the original \textsc{Poseidon} paper \citep{herde2024poseidon}.
All solutions are computed using the finite element method with bilinear quadrilateral elements on a structured rectangular mesh and $2\times2$ Gauss quadrature for numerical integration, adhering to the high-fidelity numerical standards of the original \textsc{Poseidon} benchmarks. By maintaining identical datasets and test splits across all model variants, we ensure that observed performance differences reflect genuine algorithmic improvements rather than data-related artifacts.

\paragraph{Physics-informed Training.}
We also investigate a training strategy that augments the labeled dataset with unlabeled source terms sampled on-the-fly during training. For each batch, we generate additional source functions $f$ from the extrapolation dataset without computing the corresponding solutions $\hat{u}$ and evaluate them using only the physics-informed loss. For instance, at each training step, the input functions in a batch are randomly selected source types (e.g., sinusoidal waves) with randomly sampled physical parameters (e.g., number of waves). This enables the model to continuously learn from a broader distribution of inputs while maintaining physical consistency, without the computational cost of generating large amounts of labeled training pairs. 

\subsection{Models}
We compare fine-tuning of pre-trained \textsc{Poseidon} foundation models \citep{herde2024poseidon} against FNO operators \citep{li2020fourier} trained from scratch on the Poisson equation downstream task. Our experimental design evaluates five configurations (Table \ref{tab:comparison}): \textsc{Poseidon} models fine-tuned with supervised data loss, physics-informed loss, or a hybrid approach that follows on-the-fly physics supervision (cf.\ Section~\ref{ssc:datasets}), and FNO models trained from scratch with either supervised data loss or physics-informed loss. For pure physics-informed training, we use the same input samples as for data-driven training, but without any labeled solutions. 
This enables systematic comparison of foundation model adaptation versus training from scratch, as well as data-driven versus physics-informed versus hybrid training strategies.

\begin{table}[ht]
\centering
\caption{Configurations and computational characteristics of pre-trained foundation models and operator learning models trained from scratch. Number of labeled training samples $M$: 4096. Batch size 40.}
\label{tab:comparison}
\begin{tabular*}{\textwidth}{@{\extracolsep{\fill}}lcc@{}}
\multirow{1}{*}{} & \textbf{\textsc{Poseidon}-T} & \textbf{FNO}  \\
\midrule
\makecell[l]{Learning \\ paradigm}  & \makecell[l]{
    - Data-driven \citep{herde2024poseidon} \\ 
    - Physics-informed (this work) \\
    - Hybrid (this work)
} &  \makecell[l]{- Data-driven \citep{li2020fourier} \\ - Physics-informed \citep{medvedev2025physics} \\  \\ }\\
\midrule
Training time & 0 h 37 min & 1 h 1 min\\
\midrule
Inference time & 1.6ms & 1.7ms\\
\midrule
\makecell[l]{Parameters} &  21M & 21M \\
\midrule
\makecell[l]{GFLOPs} & 0.3e3 & 1.7e3 \\
\end{tabular*}
\end{table}
Critically, the \textsc{Poseidon} (physics) and \textsc{Poseidon} (hybrid) models represent our main contribution: the first application of physics-informed fine-tuning to data-based PDE foundation models. In particular, \textsc{Poseidon} (physics) enables zero-shot adaptation with respect to labeled solution data for new PDE tasks, relying solely on physics-based supervision. In contrast, the hybrid variant additionally uses the same amount of labeled data as the purely data-driven baseline and optimizes a joint objective consisting of data and physics losses, weighted equally. We employ the T-configuration of \textsc{Poseidon} with 21M parameters. While larger variants (B with 158M and L with 629M parameters) enable more effective knowledge transfer from pre-training and substantially reduce task-specific data requirements, we deliberately select the T-configuration for its comparability in typical size to standard neural operators models. Moreover, for this specific Poisson problem, we did not observe significant performance improvements when using the larger B or L variants, suggesting that the T-configuration provides a suitable balance of efficiency and accuracy for our experiments. This allows us to investigate its predictive capabilities and explore whether physics-informed fine-tuning can push its performance beyond what data-driven training alone achieves for models of this scale. \par
To ensure fair comparison, we maintain identical experimental settings across all configurations, including batch size, learning rate scheduling, number of epochs, and number of training and test samples. The only architectural modification is a reduction in FNO model size compared to~\cite{herde2024poseidon} to approximately 21M trainable parameters, matching the parameter count of \textsc{Poseidon}-T and enabling a parameter-matched comparison between the two architectures. 
All training, fine-tuning, and inference experiments were conducted on an NVIDIA H200 GPU with 141 GB of VRAM. Both \textsc{Poseidon} and FNO maintain comparable inference latency of milliseconds, making them equally suitable for time-critical applications. 
The reduction in FLOPs $4\times$ can be explained by the architectural differences between models. The computational overhead of larger foundation models is thus amortized during the pre-training phase, while downstream fine-tuning remains efficient. Moreover, incorporating physics-informed objectives introduces only minor additional computational overhead: while derivative evaluations typically constitute the most expensive component of the PDE residual computation, the overall cost of residual evaluation remains small compared to the forward pass.

\section{Results and Discussion}
\begin{figure}[ht]
\begin{center}
\includegraphics[width=0.9\textwidth]{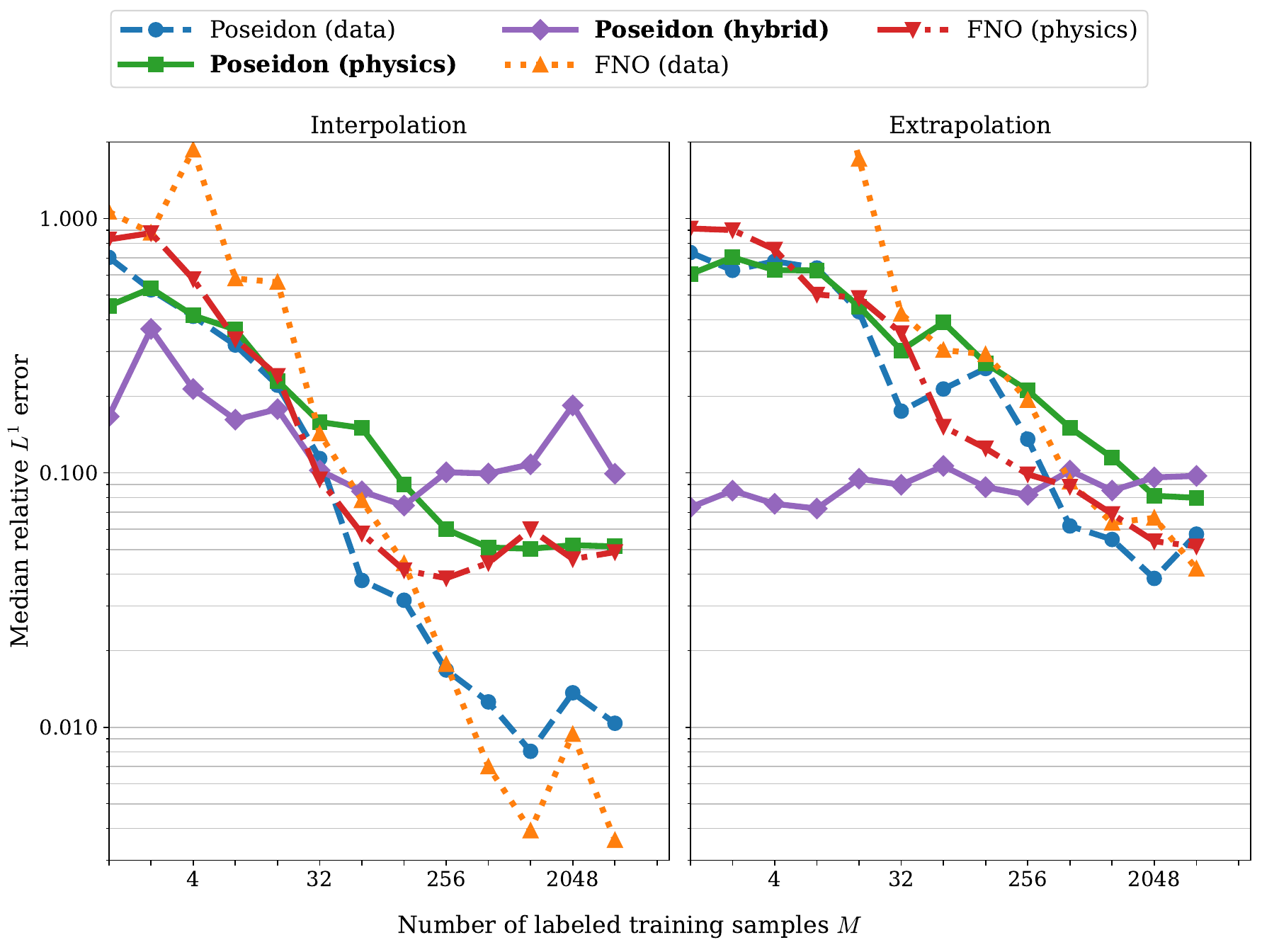}
\end{center}
\caption{Number of labeled training samples $M$ vs.\ median relative $L^1$ error of the predicted solution compared to the ground-truth solution across 240 test samples from the training distribution (left) and across 50 out-of-distribution test samples (right).}
\label{fig:figure3}
\end{figure}

\paragraph{Interpolation Performance.}
We begin by evaluating the interpolation capabilities of our physics-informed fine-tuning framework on test samples drawn from the same distribution as the training data. Figure~\ref{fig:figure3} (left panel) presents data scaling curves that reveal the relationship between training set size $M$ and median relative $L^1$ error across 240 test samples with Gaussian sources. \par

Most models achieve strong interpolation performance: error decreases consistently with increasing labeled training samples $M$. In data-scarce settings ($M<16$), fine-tuned foundation models consistently outperform FNO baselines, supporting the hypothesis that pre-training on diverse physical systems yields transferable representations that substantially lower the data requirements for new downstream tasks with unseen PDEs.\par
Notably, physics-informed FNO demonstrates better performance than data-driven FNO in the low- to mid-data regime. The physical constraints provide additional regularization that prevents overfitting and guides the model toward physically consistent solutions. Only after obtaining sufficient labeled training samples ($M \geq 128$) does the data-driven FNO begin to outperform its physics-informed counterpart, as the abundant labeled data compensates for the lack of physics-based inductive bias.\par
While data-driven and physics-informed fine-tuning approaches demonstrate a competitive accuracy (for $M<16$), the hybrid approach shows superior prediction performance, driven by its exposure to a broader diversity of source shapes through on-the-fly augmentation of unlabeled samples. A critical observation emerges: \textsc{Poseidon} (hybrid) can learn from a single labeled sample ($M=1$) and acquire the rest of the necessary knowledge via physics-informed fine-tuning, while the model with data-driven fine-tuning requires at least 32 paired input-output labeled samples to reach comparable performance levels. \par
\begin{figure}[ht]
\begin{center}
\includegraphics[width=\textwidth]{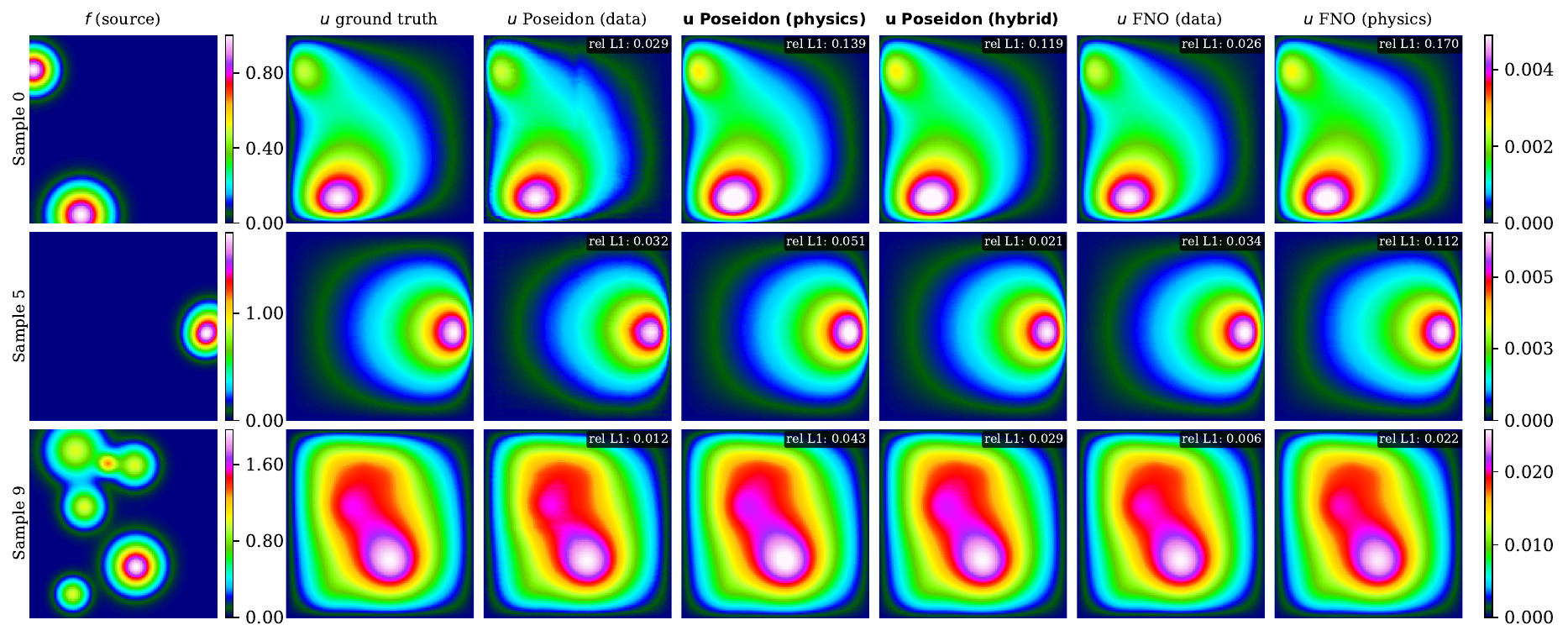}
\end{center}
\caption{Interpolation performance: qualitative comparison of predicted solutions for test samples from the training distribution. Number of labeled training samples $M$: 2048. Batch size: 40.}
\label{fig:figure2}
\end{figure}

As expected, data-driven approaches gain strength with larger training sets $M$. For instance, at $M=2048$ (Figure \ref{fig:figure2}), models incorporating data supervision achieve a substantially lower median relative $L^1$ error than physics-informed counterparts. Although the data-driven approach achieves the lowest pixel-wise error, evaluation of the governing PDE residual using a Laplacian finite-difference stencil reveals that the model primarily learns an image-to-image mapping rather than the underlying physics. In contrast, physics-informed fine-tuning yields solutions that better satisfy the finite-difference stencil and, consequently, the governing equations. Additional results on physics consistency are reported in the Appendix \ref{sec:appendixA}.

\begin{figure}[ht]
\begin{center}
\includegraphics[width=\textwidth]{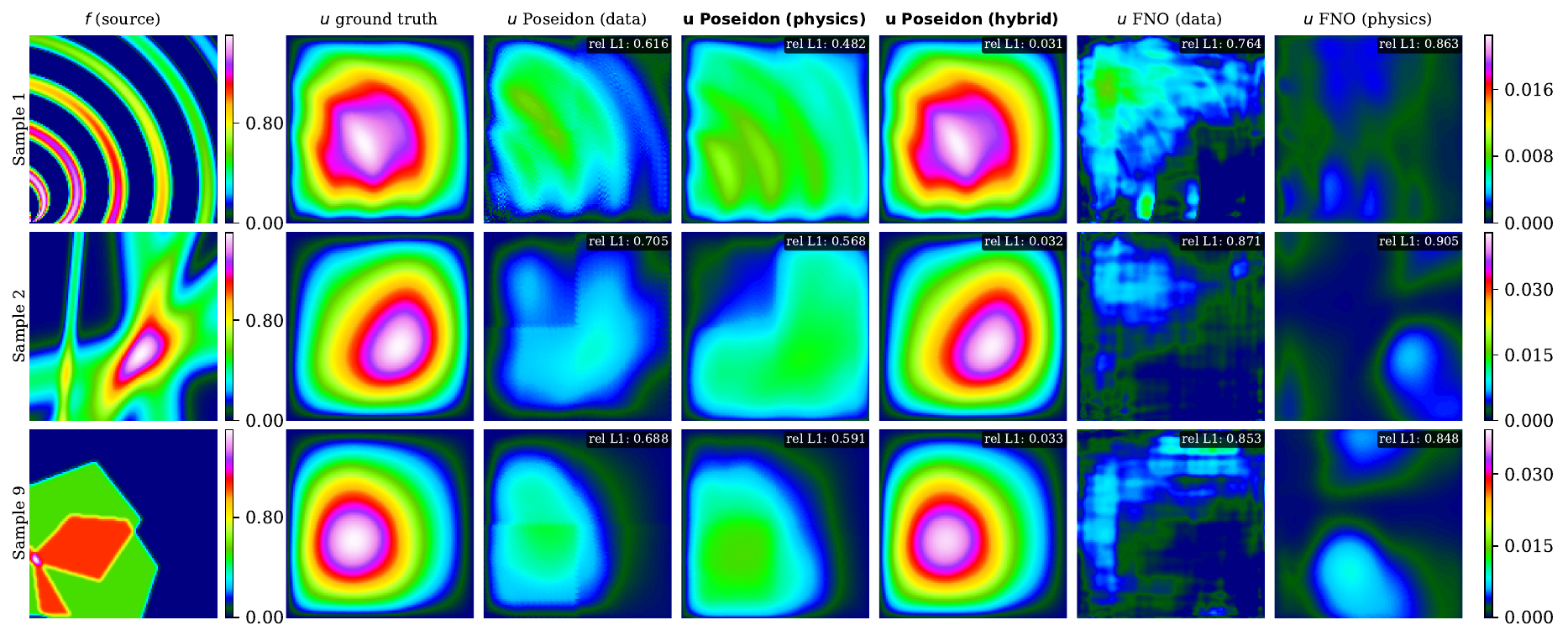}
\end{center}
\caption{ Extrapolation performance: qualitative comparison of model predictions on out-of-distribution test samples. Number of labeled training samples $M$: 1. Batch size: 1.}
\label{fig:figure4}
\end{figure}

\paragraph{Extrapolation Performance.}
Comparing the data-driven approaches, we observe the advantage of parameter transfer underlying the foundation model in the low-data regime, as demonstrated in Figure~\ref{fig:figure3} (right panel). The data-driven FNO struggles significantly when predicting from unseen extreme source shapes, resulting in degraded performance. In contrast, physics-informed FNO exhibits greater stability due to the superior extrapolation capabilities of physics-informed machine learning, where physical constraints act as `guardrails' that enforce physically plausible predictions. Notably, however, the \textsc{Poseidon} (hybrid) shows a stagnation and limited improvement in extrapolation performance as $M$ increases, likely because additional labeled samples come exclusively from the interpolation distribution (Gaussian sources), causing in-distribution samples to become overrepresented and diluting the benefit of physics-informed on-the-fly augmentation from diverse source shapes. As expected, increasing the number of labeled training samples improves the performance of all data-driven models in both interpolation and extrapolation settings. Visually, models embedding physics as regularization loss terms consistently produce smoother predictions compared to the partly noisy outputs of purely data-driven approaches, cf.\ Figure~\ref{fig:figure4}. \par
The effectiveness of \textsc{Poseidon} (hybrid) becomes particularly pronounced in the extrapolation task due to its ability to sample arbitrary unlabeled training samples. While labeled supervision is restricted to interpolation samples including Gaussian sources, the hybrid model is effectively trained under an expanded distribution of source geometries.
This explains why the extrapolation error of \textsc{Poseidon} (hybrid) can be lower than its interpolation error (Figure~\ref{fig:figure3}), as the augmented training distribution better aligns with the extrapolation test set while interpolation performance remains constrained by the limited labeled samples. \par

By incorporating more diverse source shapes during training, it achieves substantially better performance and lower error, especially in the low-data regime. For example, qualitative extrapolation results on three test samples with unseen extreme source shapes are shown in Figure \ref{fig:figure4}. All models except \textsc{Poseidon} (hybrid) exhibit poor prediction performance when trained on a single labeled source-solution pair ($M=1$).
For this specific setup, the \textsc{Poseidon} model, fine-tuned in a physics-informed manner with on-the-fly sample augmentation, exhibits  the lowest errors, demonstrating robust extrapolation through physical constraints. Critically, this physics-informed approach enables training without requiring ground truth solutions, a capability unavailable to purely data-driven models, which fundamentally depend on paired source-solution data. This highlights the dual advantage of physics-informed learning: superior handling of distribution shift to extreme, out-of-distribution source geometries, and the ability to augment the training data from physical laws alone.\par

In general, foundation models offer compelling practical advantages over training FNO from scratch. In low-data regimes, fine-tuned foundation models achieve comparable or superior accuracy to FNO while requiring significantly less training time and computational resources (Table \ref{tab:comparison}). This efficiency gain stems from leveraging pre-trained representations that already capture general physical priors, allowing rapid adaptation to new tasks without the computational burden of training large operator networks from scratch for each downstream application.

\section{Conclusion}
This work introduces, to the best of our knowledge, the first systematic investigation of physics-informed fine-tuning for data-based PDE foundation models. By incorporating PDE residuals and boundary conditions directly into the fine-tuning objective, we enable effective adaptation to new physical domains. The proposed approach offers an alternative to providing large amounts of known solutions: the model learns the underlying PDE through residual minimization. Our experimental results on the Poisson and Helmholtz PDE downstream task demonstrate that physics-informed fine-tuning achieves competitive prediction performance when only minimal training data are available. Critically, these results were achieved without any labeled solution data, validating the core premise that physics-informed objectives can provide effective supervision for foundation model adaptation. \par
Beyond a straightforward application of existing techniques, our framework involves non-trivial methodological design choices, particularly the hybrid strategy with on-the-fly generation of unlabeled input functions. This approach requires carefully coupling synthetic sample generation, PDE-constrained residual minimization, and optional data-driven supervision within a unified optimization framework. We demonstrate that hybrid fine-tuning, which combines physics-informed and data-driven objectives, achieves superior extrapolation and generalization to out-of-distribution scenarios. When experimental or observational data is prohibitively expensive or impossible to obtain, purely physics-informed fine-tuning provides supervision with zero observational data, if governing physics is known. By generating synthetic input functions on the fly and enforcing PDE constraints, models can self-supervise without requiring ground-truth solutions from numerical solvers or physical experiments. This capability could prove transformative for scientific domains where data acquisition is the primary bottleneck.\par
Several limitations of our current study point toward promising research directions. First, the relatively simple steady-state Poisson and Helmholtz equations may not fully expose the benefits of physics-informed learning. Future work will therefore investigate more complex settings, including time-dependent PDEs, coupled multiphysics systems, and problems with sharp gradients or discontinuities, where the advantages of physics-informed fine-tuning may become more pronounced. Second, foundation model pre-training on observational data may introduce biases that limit the impact of physics-based fine-tuning. One potential direction is physics-informed pre-training, which could combine the representational capacity of foundation models with the strong generalization properties of physics-informed learning. In addition, evaluating the proposed approach across a broader range of PDE foundation models would help assess its generality and robustness.\par
 Finally, it is worth noting that all experiments in this work used the original hyperparameters from the foundation model \citep{herde2024poseidon} without physics-specific tuning, ensuring fair comparison but potentially leaving performance gains on the table. Future work will investigate optimization strategies tailored to physics-informed fine-tuning, explore hybrid loss formulations that optimally balance data-driven and physics-informed terms, and conduct comprehensive benchmarking across the full spectrum of PDEs represented in the foundation model pre-training datasets. Ultimately, this work takes a critical step toward trustworthy, interpretable, and data-efficient machine learning surrogates for computational science and engineering, demonstrating that physics-informed learning and foundation models can be synergistically combined to advance scientific machine learning.

\subsubsection*{Acknowledgments}
This work was supported by the Fraunhofer Internal Programs under Grant No. PREPARE 40-08394.

\bibliography{iclr2026_conference}
\bibliographystyle{iclr2026_conference}

\appendix
\section{Helmholtz PDE}
\label{sec:appendixB}

To further demonstrate the significance of the proposed work across different PDE types, 
we additionally consider a two-dimensional Helmholtz equation, which describes wave 
propagation phenomena in the frequency domain. Specifically, we study the following variant:
\begin{equation}
    -\Delta u - \omega^2 a(x,y)\, u = 0, \quad \text{on } D := (0,1)^2,
    \label{eq:helmholtz}
\end{equation}
subject to Dirichlet boundary conditions $u = b$ on $\partial D$, where $\omega = 5\pi/2$ 
denotes the frequency, $a(x,y)$ is a spatially varying function encoding the properties 
of the propagation medium, and $b$ is a prescribed constant boundary value sampled from a uniform distribution $b \sim \mathcal{U}_{[0.25, 0.5]}$.\par
\begin{figure}[ht]
\begin{center}
\includegraphics[width=1\textwidth]{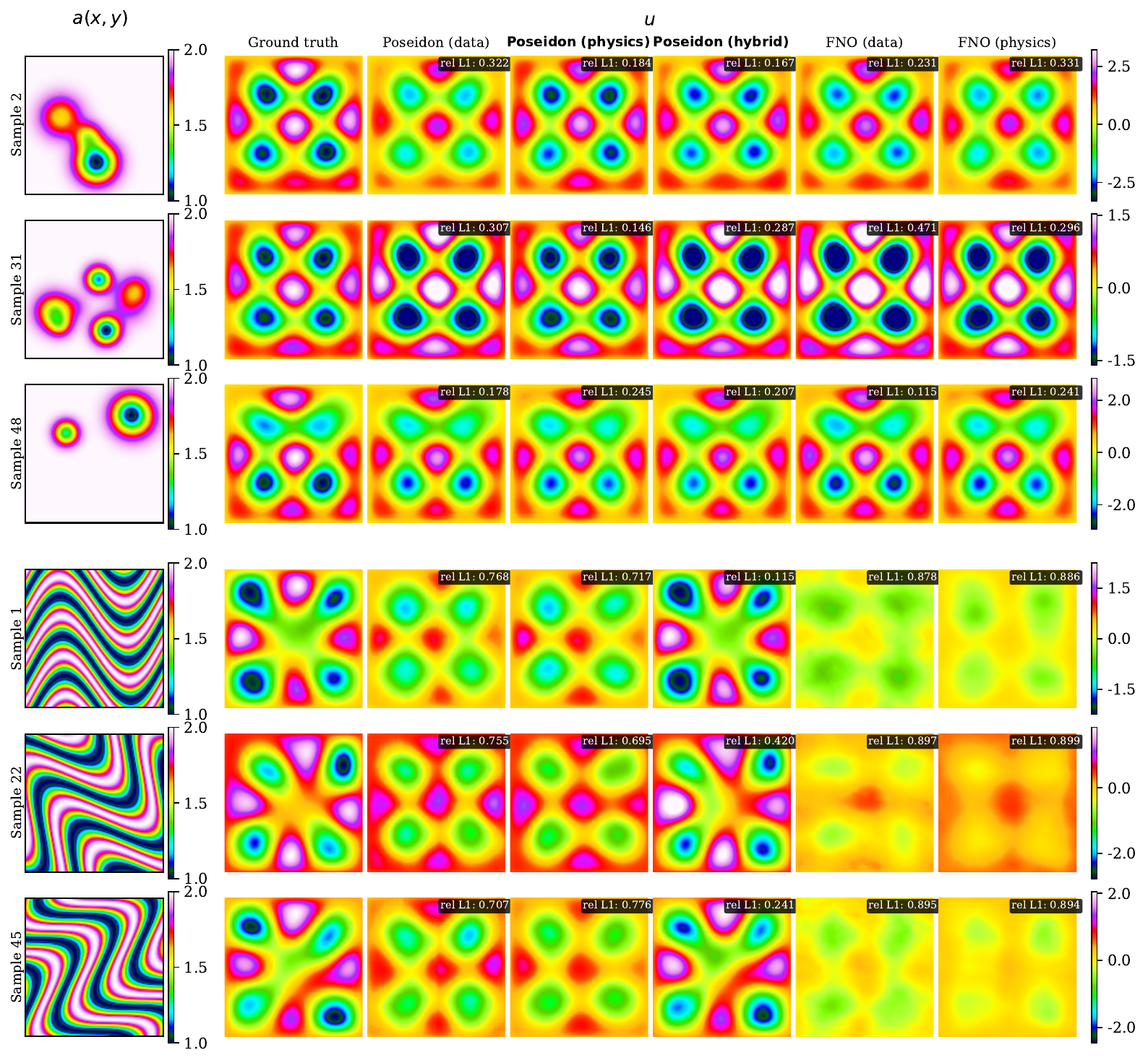}
\end{center}
\caption{Qualitative comparison of model predictions for the Helmholtz equation. Top three rows: interpolation performance for test samples from the training distributions ($a(x,y)$ composed of Gaussians). Bottom three rows: extrapolation performance for out-of-distribution test samples ($a(x,y)$ composed of wavy stripes). Number of labeled training samples $M$: 2048. Batch size: 40.}
\label{fig:figure5}
\end{figure}

 
During training, all models observe exclusively coefficient fields $a(x,y)$ constructed as a superposition of a random number of Gaussian components (Figure \ref{fig:figure5}, top three rows). 

In the extrapolation regime, the coefficient field $a(x,y)$ is replaced entirely by qualitatively different media, specifically rotated wavy stripes with randomly sampled orientation, stripe count, and waviness amplitude (Figure \ref{fig:figure5}, bottom three rows). This setup explicitly tests the model’s ability to generalize beyond the structural complexity observed during training.
\appendix
\setcounter{section}{1}
\section{Additional Results on Physics Consistency}
\label{sec:appendixA}
In this appendix, we provide additional quantitative results evaluating physics consistency of the predicted solutions with respect to the governing Poisson equation residual (Equation \ref{eq:poisson}). Specifically, we compute the median relative $L^1$ error of the resulting residuals using a second-order finite-difference Laplacian stencil.\par
\begin{figure}[H]
\begin{center}
\includegraphics[width=0.9\textwidth]{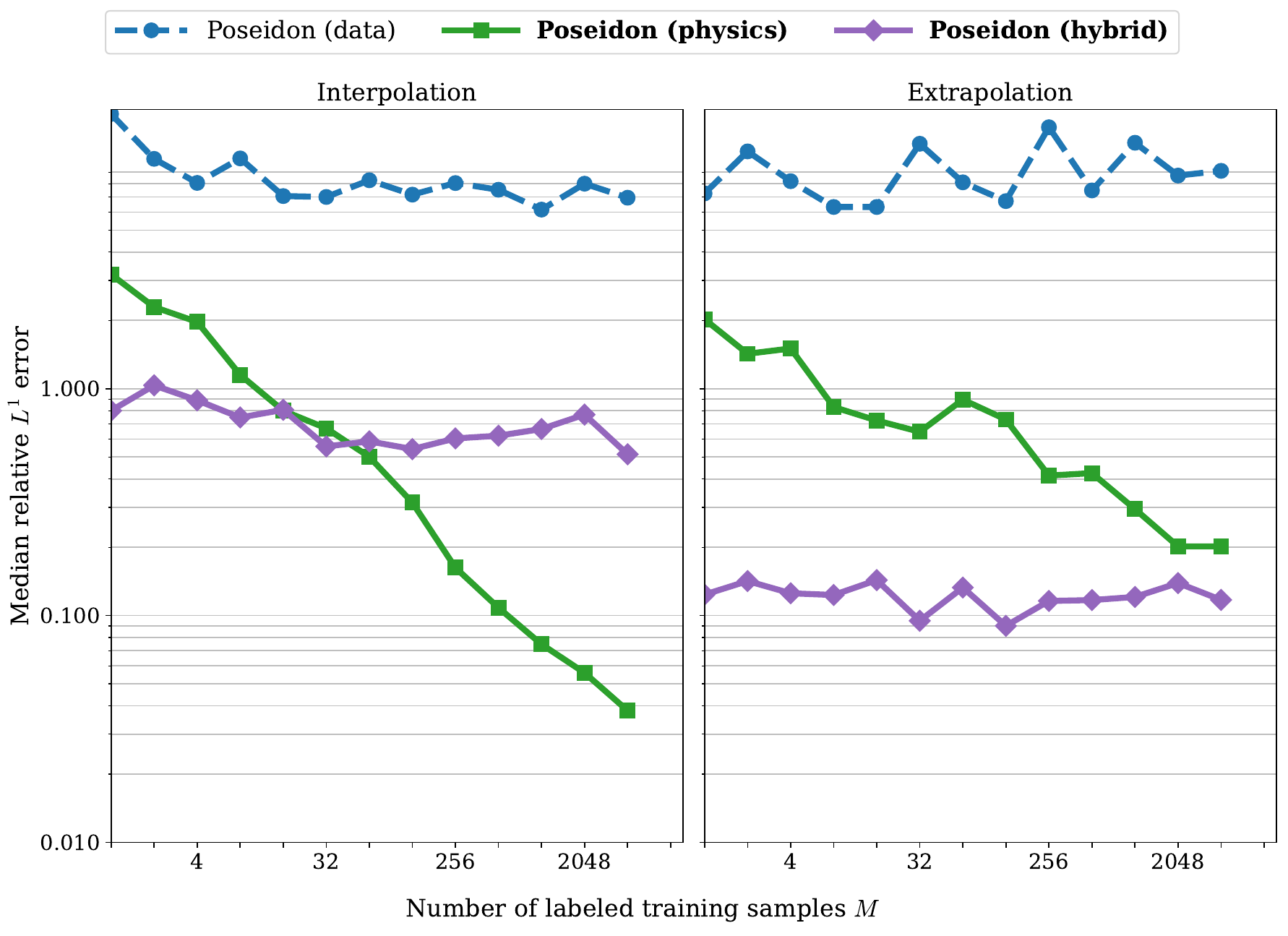}
\end{center}
\caption{Number of labeled training samples $M$ vs.\ median relative $L^1$ error of the Poisson equation residual across 240 test samples from the training distribution (left) and across 50 out-of-distribution test samples (right).}
\label{fig:figure6}
\end{figure}

While data-driven fine-tuning achieves low pixel-wise error (Figure \ref{fig:figure2}), its high governing PDE residuals suggest the model prioritizes image-to-image mappings over the underlying physical laws. In contrast, the physics-informed approach shows a dramatic improvement in consistency as the number of training samples $M$ increases, particularly within the interpolation regime shown in Figure \ref{fig:figure6} (left panel). The physics-only model exhibits a slight performance degradation when the test distribution shifts from Gaussian sources to more diverse source geometries (Figure \ref{fig:figure6} (right panel)). Ultimately, these results demonstrate that physics-informed fine-tuning forces the model to respect the governing equations, leading to solutions that are more physically consistent than those produced by purely data-driven supervision.

\end{document}